\documentclass[runningheads]{llncs}

\usepackage{eccv}

\usepackage{eccvabbrv}

\usepackage{graphicx}
\usepackage{booktabs}

\usepackage[accsupp]{axessibility}  

\usepackage{hyperref}

\usepackage{orcidlink}

\begin{document}

\title{AnchorSplat: Fast and Structure Consistent Detail Synthesis for Gaussian Splatting} 

\titlerunning{AnchorSplat}

\author{Dexu Zhu\inst{1}\textsuperscript{*} \and
Jiangnan Shao\inst{1,2,3}\textsuperscript{*} \and
Xiaofeng Wang\inst{2,4} \and
Junxian Duan\inst{1} \and \\
Jie Cao\inst{1}\textsuperscript{\dag} \and
Zheng Zhu\inst{2} \and
Huaibo Huang\inst{1}
}

\authorrunning{D. Zhu et al.}

\institute{\quad $^1$ MAIS\&NLPR, CASIA, Beijing, China \\ \quad $^2$ GigaAI, Beijing, China \\ \quad $^3$ ShanghaiTech University, Shanghai, China \\ \quad $^4$ Tsinghua University, Beijing, China
 \\
\email{dexu.zhu@cripac.ia.ac.cn}
}

\maketitle

\begingroup
\renewcommand{\thefootnote}{*}
\footnotetext{Equal contribution.}

\renewcommand{\thefootnote}{\dag}
\footnotetext{Corresponding author.}
\endgroup


\begin{abstract}
3D Gaussian Splatting (3DGS) has emerged as a powerful representation for high-fidelity rendering. However, existing assets often suffer from quality bottlenecks such as missing details and texture noise. Prior attempts to enhance these assets via 2D image processing introduce multi-view inconsistencies and high computational costs. In this paper, we propose a novel 3D-native refinement paradigm named \textbf{AnchorSplat}. AnchorSplat is an end-to-end deep network operating directly on 3D structures, avoiding the expensive optimization overhead of traditional 3D-2D-3D pipelines. Crucially, AnchorSplat is a strictly source-free solution requiring no original multi-view images. Central to the proposed method is the Point Anchor Mechanism, which enforces geometric consistency via local offset constraints, mitigating ill-posed mapping and gradient confounding. Furthermore, AnchorSplat replaces iterative densification with a single-pass multiplication mechanism. To facilitate research, we construct 3DGS-SR, the first large-scale benchmark for this task. Experiments demonstrate state-of-the-art results on the 3DGS-SR dataset, with throughput \textbf{up to $\mathbf{10^5}$ times faster} than optimization methods. Notably, AnchorSplat exhibits robust \textbf{zero-shot generalization} across diverse data distributions, including generative model outputs and real-world scans. The repository is available at: \url{https://github.com/zhude233/AnchorSplat}
\end{abstract}

\section{Introduction}
\label{sec:intro}

3D Gaussian Splatting (3DGS)~\cite{kerbl20233d, wu20244d, huang20242d, wang2024hallo3d}, despite its success in high-fidelity rendering, suffers from significant quality degradation, including sparse geometry and missing details, when reconstructed from Low-Resolution (LR) inputs. To overcome this limitation, the task of 3D Super-Resolution (3DSR) has emerged, which aims to enhance these low-quality assets into High-Resolution (HR), detail-rich counterparts. Addressing this quality degradation is critical, as these deficiencies severely restrict applicability in detail-demanding applications.



Prior works~\cite{feng2024srgs, ko2025sequence, wan2025s2gaussian} at 3DSR primarily rely on 2D image-level post-processing. This paradigm does not directly leverage the structured 3D geometric information, leading to critical limitations. First, processing isolated 2D images fails to enforce strict 3D consistency, introducing multi-view artifacts. Second, the required 2D Super-Resolution (2DSR) and subsequent 3D re-optimization incur significant computational overhead. Furthermore, these pipelines are inherently source-dependent, requiring original multi-view images to supervise the re-optimization. This renders them unusable in highly practical source-free scenarios where only the raw 3D asset is available, such as enhancing the unconstrained outputs from recent feed-forward 3D generative models.

The logical alternative to these 2D-centric methods is a 3D-native paradigm. However, while recent works~\cite{chen2024splatformer, zhang2025gap, zhou2024diffgs, yan20243dsceneeditor, xu2025resplat} do manipulate 3DGS structures, they focus on different tasks, such as generation or view generalization, and do not address the specific problem of enhancing existing, low-quality assets.

\begin{figure}[t!]
\centerline{\includegraphics[width=\columnwidth]{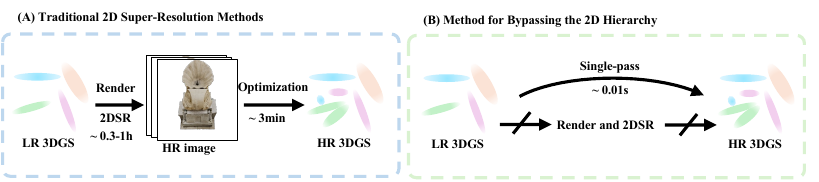}}
\caption{
Comparison of 3DSR paradigms. (A) depicts the conventional 2D-centric pipeline: the process requires rendering the asset to LR images, applying 2DSR, and then performing 3D reconstruction. This multi-step process is computationally expensive. (B) shows our novel 3D native paradigm: by directly processing the 3D input, we bypass the costly intermediate 2DSR step, thus achieving significantly higher throughput and full utilization of 3D geometry.
}
\label{fig:pipeline_compare}
\end{figure}

Consequently, a critical research gap persists for a solution that is both 3D native and designed for enhancement. Motivated by this specific gap, we propose AnchorSplat, a framework that rapidly enhances 3DGS asset quality. We pursue a novel high-efficiency technical path: operating directly on the asset structure in the 3D space.
This 3D native approach, shown in \cref{fig:pipeline_compare} (B), is intended to avoid the multi-view inconsistencies and expensive optimization overhead associated with the conventional 2D-centric processing pipelines, which are depicted in \cref{fig:pipeline_compare} (A). Our architecture is designed as an efficient, single-pass feed-forward network. It first processes the low-quality 3DGS asset as an attribute-rich 3D point cloud. We then employ an efficient point cloud encoder to aggregate local context, extracting a high-dimensional feature descriptor for each individual input Gaussian primitive. 
This design, which relies on decoding these per-point features, faces two fundamental challenges.

First is the ill-posed mapping problem stemming from the unstructured, point cloud nature of 3DGS. Using a per-point feature to generate an unconstrained new primitive leads to severe gradient confounding, as it can receive conflicting signals from arbitrary views and locations. To resolve this, we introduce the Point Anchor Mechanism. This mechanism leverages the existing coarse geometry of the input as a reliable foundation. It then uses the feature extracted from each anchor to generate new detail primitives only within the immediate vicinity of that anchor, enforcing a strict local constraint that preserves the valid structure and mitigates the mapping ambiguity.

Second, this single-pass encoder design is incompatible with the iteration-based densification of native 3DGS, namely cloning and splitting. To replace this, AnchorSplat uses the same per-point features to drive a controlled multiplicative generation mechanism, named Equivalent Densification Mechanism. This mechanism achieves equivalent density control. It substitutes for cloning by multiplicatively generating new primitives, and substitutes for pruning by learning vanishing opacity.

Experimental results demonstrate that AnchorSplat achieves significant improvements in both performance and efficiency. We evaluate our method on the newly constructed 3DGS-SR benchmark, where it achieves state-of-the-art fidelity while demonstrating a processing throughput increase of up to $10^5$ times over conventional optimization methods. Crucially, AnchorSplat exhibits remarkable zero-shot generalization capabilities in highly practical, source-free scenarios. Without any fine-tuning or access to original multi-view images, our model robustly enhances unconstrained assets generated by 3D large generative models as well as noisy real-world object captures. This confirms that AnchorSplat learns dataset-agnostic geometric priors, making it a universal and highly efficient detail enhancer for the broader 3D asset ecosystem.

Overall, our contributions can be summarized as follows:
\begin{itemize}
    \item We propose AnchorSplat, a novel 3D-native, feed-forward refinement framework that achieves a $10^5\times$ speedup in processing throughput compared to traditional optimization-based enhancement methods.
    \item We introduce the Point Anchor Mechanism to resolve the gradient confounding inherent in unconstrained 3D prediction, and an Equivalent Densification Mechanism as a single-pass substitute for iterative 3DGS optimization.
    \item We construct 3DGS-SR, the first large-scale benchmark specifically designed for 3DGS asset enhancement. AnchorSplat achieves state-of-the-art results on this benchmark and demonstrates strong zero-shot generalization to unconstrained AIGC assets and real-world captures under strict source-free conditions.
\end{itemize}

\section{Related Work}

\subsection{Novel View Synthesis}

Novel View Synthesis (NVS) aims to synthesize photorealistic images from unseen viewpoints given a set of input images. Neural Radiance Fields (NeRF)~\cite{mildenhall2021nerf, lin2024dynamic} learn continuous implicit scene representations with MLPs and render them via volumetric rendering. In contrast, 3D Gaussian Splatting (3DGS)~\cite{kerbl20233d} represents scenes explicitly with anisotropic Gaussians and enables real-time rendering through differentiable rasterization while maintaining high visual quality. Despite these advances, both implicit and explicit representations remain highly dependent on input quality. When reconstructed from sparse or low-resolution images, 3DGS assets often suffer from sparse geometry, missing details, and texture noise.

\subsection{3D Super-Resolution}

Existing 3D super-resolution methods can be broadly categorized into reconstruction from scratch and asset enhancement. Reconstruction-based methods directly optimize high-resolution 3D representations from low-resolution multi-view inputs~\cite{ko2025sequence, feng2024srgs, wan2025s2gaussian, zeng2025arbitrary}, often leveraging 2D image priors~\cite{huang2025infrared, kim2016deeply, li2023self}. For example, SRGS~\cite{feng2024srgs} uses SISR models~\cite{dong2015image, liang2021swinir, saharia2022image} to generate high-resolution pseudo-labels with a sub-pixel constraint, while SuperGS~\cite{shen2024supergaussian} further incorporates variational residual features. Other methods exploit VSR priors~\cite{xu2025videogigagan, liu2025ultravsr, xiao2025event, zheng2025efficient}; for instance, Sequence Matters~\cite{ko2025sequence} treats multi-view images as video-like sequences~\cite{duan2025dual, zhu2025mtsd} to improve temporal and spatial coherence. However, such 2D-to-3D strategies remain limited by the mismatch between 2D enhancement priors and the multi-view consistency required by 3D scenes. Asset enhancement methods instead start from an existing low-quality 3D asset and upgrade it to a higher-fidelity representation. Existing approaches~\cite{han2024super, wu2024rafe} typically follow a 3D-2D-3D pipeline: rendering the coarse 3DGS asset into image sequences, enhancing them with 2D or video super-resolution models, and re-optimizing the 3D representation. Although this pipeline can improve visual details, it introduces additional rendering and reconstruction stages, increasing computational cost and potentially causing multi-view artifacts. In contrast, our method performs enhancement entirely in the 3D domain. It directly takes a low-resolution 3DGS asset as input and outputs a refined Gaussian representation without intermediate 2D conversion or re-optimization. This design avoids 2D-3D domain mismatch, better preserves structural consistency, and enables substantially higher throughput.

\subsection{3D Point Cloud}

\begin{figure*}[tbp]
\centerline{\includegraphics[width=\textwidth]{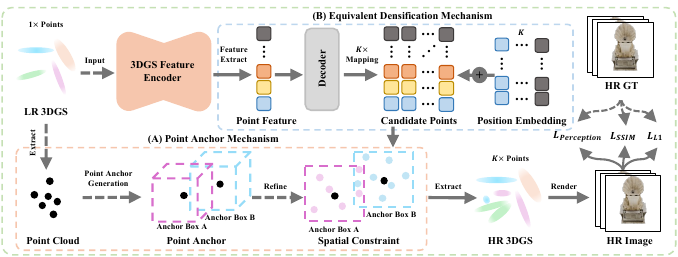}}
\caption{
Framework of AnchorSplat. Our model takes a low-quality 3DGS asset as input and first encodes its non-positional attributes into per-point features. (A) The Point Anchor Mechanism imposes a local geometric constraint by defining an anchor box around each input primitive. (B) The 3DGS Decoder then multiplicatively generates $K$ detail-enhanced primitives strictly within this localized anchor space, utilizing a learnable position embedding. The final high-quality 3DGS asset is rendered and supervised end-to-end against high-resolution ground truth, ensuring strong geometric consistency and fidelity improvement.
}
\label{fig:model_pipeline}
\end{figure*}

Point clouds~\cite{sohail2025advancing, zhang2025survey, dumic2025three, lu2026current} serve as the foundational representation for 3D geometry. Early works such as PointNet~\cite{qi2017pointnet} and PointNet++~\cite{qi2017pointnet++} introduced the idea of directly processing point clouds via MLP and local aggregation. In recent years, Transformer-based models~\cite{zhao2021point, wu2022point, wu2024point, fan2024large, zhu2024advancements} have achieved state-of-the-art performance across various 3D tasks.

Our work builds upon the profound insights derived from these point cloud networks. At its core, 3DGS is an attribute-rich, unstructured 3D point cloud. PTv3~\cite{wu2024point, han2024mamba3d}, with its simplified architecture and efficient attention mechanism, is highly suitable for modeling 3DGS primitives. While most prior methods focus on tasks like classification or segmentation~\cite{liang2024pointmamba, han2024mamba3d, wang2024unlearnable}, our work targets 3DSR. We utilize PTv3 to effectively extract local and global geometric context in a purely 3D manner, which provides the crucial semantic basis for the Point Anchor Mechanism.

\section{Method}

\subsection{Overview}

Existing 3D super-resolution methods primarily achieve high-resolution results indirectly through 2D image enhancement and 3D optimization. This approach struggles to guarantee robust multi-view 3D consistency and cannot directly exploit the explicit geometric structure of 3DGS assets. To overcome these limitations, we propose AnchorSplat.

AnchorSplat is designed to transform a set of low-quality 3DGS assets $\{\mathcal{G}_{\text{low}}\}_{i=1}^{N}$ into detail-rich, high-fidelity assets $\{\mathcal{G}_{\text{high}}\}_{i=1}^{N}$. Our task is formally defined as:
\begin{equation}
\mathcal{G}_{\text{high}} = \mathcal{F}_{\text{AnchorSplat}}(\mathcal{G}_{\text{low}}).
\end{equation}
The core components of $\mathcal{F}_{\text{AnchorSplat}}$ include: (1) a 3DGS Feature Encoder, (2) the Point Anchor Mechanism, and (3) the Equivalent Densification Mechanism. The overall framework of AnchorSplat is illustrated in \cref{fig:model_pipeline}.

\subsection{3DGS Feature Encoder}

We treat the low-quality $3\text{DGS}$ asset $\mathcal{G}_{\text{low}}$ as an attribute-rich $3\text{D}$ point cloud. The input is a set of $n$ Gaussian primitives $\mathcal{G}_{\text{low}} = \{(\boldsymbol{\mu}_j, \mathbf{c}_j, \mathbf{s}_j, \mathbf{R}_j, \alpha_j)\}_{j=1}^{n}$, where $\boldsymbol{\mu}_j$ denotes the $3\text{D}$ position, and the remaining terms $\left(\mathbf{c}_j, \mathbf{s}_j, \mathbf{R}_j, \alpha_j\right)$ represent the rendering attributes. To extract features with local context awareness from this sparse, unstructured representation, we first concatenate all non-positional attributes into a high-dimensional initial attribute feature vector $\mathbf{v}_{j} \in \mathbb{R}^M$:
\begin{equation}
\mathbf{v}_{j} = \text{Concat}(\mathbf{c}_j, \text{Vec}(\mathbf{s}_j), \text{Vec}(\mathbf{R}_j), \alpha_j) .
\end{equation}
Subsequently, we input $\boldsymbol{\mu}_j$ and $\mathbf{v}_j$ into the Point Transformer v3 (PTv3) encoder $\mathcal{E}_{\text{PTv3}}$. $\mathcal{E}_{\text{PTv3}}$ efficiently aggregates geometric and attribute information within the neighborhood, transforming the sparse input primitives into a robust high-dimensional feature descriptor $\mathbf{f}_j \in \mathbb{R}^D$:
\begin{equation}
\mathbf{f}_j = \mathcal{E}_{\text{PTv3}}(\boldsymbol{\mu}_j, \mathbf{v}_{j}).
\end{equation}
The resulting feature $\mathbf{f}_j$ provides the crucial semantic basis for the Point Anchor Mechanism.

\subsection{Point Anchor Mechanism}

The optimization objective of AnchorSplat is to minimize the total rendering loss across all views $\mathcal{V}$, defined as $ \mathcal{L}_{\text{total}} = \sum_{v \in \mathcal{V}} \mathcal{L}_v $ .
The core challenge of this process lies in the mapping from the 3D feature $\mathbf{f}_j$, extracted by the encoder as a local descriptor in the vicinity of $\boldsymbol{\mu}_j$, to the global novel primitives $g'_{j,k}$ generated by the decoder $\mathcal{D}_{\text{Gen}}$. An unconstrained decoder $\mathcal{D}_{\text{Gen}}$ results in a fundamental mapping ambiguity.

In practice, such unconstrained training leads to catastrophic convergence failure: while the network learns the coarse envelope of the scene, it ultimately decodes highly blurry geometry with chaotic colors, indicating that the precise optimization signals required for high-frequency details are completely lost.

To investigate the fundamental reason of this convergence barrier, we analyze the global gradient received by $\mathbf{f}_j$. Early in training, the decoder $\mathcal{D}_{\text{Gen}}$ is randomly initialized, and there lacks any prior spatial correspondence between $\mathbf{f}_j$ and the primitives $g'_{j,k}$ it generates. Let $\mathcal{P}_v$ be the set of all pixels in view $v$, and $\mathbf{C}_p^v$ be the rendered color of pixel $p$. According to the chain rule, the total gradient with respect to $\mathbf{f}_j$ is the summation of the rendering loss contributions from its $K$ generated primitives $g'_{j,k}$ over all views $v$ and all pixels $p \in \mathcal{P}_v$:
\begin{equation}
\label{eq:total_backward}
\frac{\partial \mathcal{L}_{\text{total}}}{\partial \mathbf{f}_j} = \sum_{v \in \mathcal{V}} \sum_{p \in \mathcal{P}_v} \frac{\partial \mathcal{L}_v}{\partial \mathbf{C}_p^v} \cdot \left( \sum_{k=1}^K \frac{\partial \mathbf{C}_p^v}{\partial g'_{j,k}} \cdot \frac{\partial g'_{j,k}}{\partial \mathbf{f}_j} \right).
\end{equation}
This mathematically reveals why this problem is ill-posed. First, for the mapping term $\frac{\partial g'_{j,k}}{\partial \mathbf{f}_j}$, an unconstrained decoder $\mathcal{D}_{\text{Gen}}$ may map $\mathbf{f}_j$ to a primitive $g'_{j,k}$ at an arbitrary global location. Second, the rendering term $\frac{\partial \mathbf{C}_p^v}{\partial g'_{j,k}}$ implies that this $g'_{j,k}$ can project onto an arbitrary pixel $p$ in an arbitrary view $v$, thus generating a gradient. Consequently, $\mathbf{f}_j$, despite being a local feature, receives conflicting and confounded gradient signals from all views and spatial locations. This is the fundamental cause for decoding blurry shapes and chaotic colors.

To resolve this mapping ambiguity and the issue of confounded gradients, we propose the Point Anchor Mechanism. This mechanism enforces the spatial correspondence by imposing an explicit 3D geometric constraint, thereby transforming the ill-posed problem into a well-posed one. We define the position of the new primitive $\boldsymbol{\mu}'_{j, k}$ as a relative offset $\Delta \boldsymbol{\mu}_{j, k}$ with respect to its anchor point $\boldsymbol{\mu}_j$:
\begin{equation}
\label{eq:anchor}
\boldsymbol{\mu}'_{j, k} = \boldsymbol{\mu}_j + \epsilon \cdot \tanh(\Delta \boldsymbol{\mu}_{j, k}).
\end{equation}
This constraint compels $\mathbf{f}_j$ to only generate new primitives $g'_{j,k}$ within a local geometric domain $\Omega_{\text{local}}(\boldsymbol{\mu}_j)$ in the vicinity of its 3D anchor $\boldsymbol{\mu}_j$. 
This constraint mathematically rewrites the gradient flow. 
Since $g'_{j,k}$ are geometrically constrained to $\Omega_{\text{local}}(\boldsymbol{\mu}_j)$, they can only project onto a local pixel set $\mathcal{P}_{\text{local}}(\boldsymbol{\mu}_j, v)$ in the 2D image of view $v$. 
Let $\mathcal{P}_{\text{local}}^j \equiv \mathcal{P}_{\text{local}}(\boldsymbol{\mu}_j, v)$ denote this view-dependent local pixel set for brevity.
For pixels $p \notin \mathcal{P}_{\text{local}}^j$ outside this local region, the primitives $g'_{j,k}$ generated from $\mathbf{f}_j$ have no contribution to $\mathbf{C}_p^v$. Consequently, the gradient $\frac{\partial \mathbf{C}_p^v}{\partial g'_{j,k}}$ is mathematically and precisely zero:
\begin{equation}
\label{eq:regular}
\text{If } p \notin \mathcal{P}_{\text{local}}^j, \quad \frac{\partial \mathbf{C}_p^v}{\partial g'_{j,k}} = 0 .
\end{equation}
Substituting this zero-gradient condition into Eq.~\eqref{eq:total_backward}, the summation over all pixels $p \in \mathcal{P}_v$ in the global gradient automatically reduces to a summation over only the local pixels $p \in \mathcal{P}_{\text{local}}$:
\begin{equation}
\label{eq:local_backward}
\frac{\partial \mathcal{L}_{\text{total}}}{\partial \mathbf{f}_j} = \sum_{v \in \mathcal{V}} \sum_{p \in \mathcal{P}_{\text{local}}^j} \frac{\partial \mathcal{L}_v}{\partial \mathbf{C}_p^v} \cdot \left( \sum_{k=1}^K \frac{\partial \mathbf{C}_p^v}{\partial g'_{j,k}} \cdot \frac{\partial g'_{j,k}}{\partial \mathbf{f}_j} \right) .
\end{equation}
This is not an approximation, but the exact mathematical result of Eq.~\eqref{eq:total_backward} under the constraint imposed by Eq.~\eqref{eq:regular}. This demonstrates that the gradients from all views are constrained to optimize the same 3D local region $\Omega_{\text{local}}$. This effectively eliminates the confounding influence of distant primitives, rendering the optimization signal for $\mathbf{f}_j$ clean and 3D-consistent. This enables the network to learn fine-grained geometric and appearance details.

\subsection{Equivalent Densification Mechanism}

With the spatial localization and gradient confounding issues addressed, the second core challenge is the generation of a sufficient density of geometric primitives, which is essential for achieving high-fidelity detail. The original 3DGS framework achieves densification through an iteration-based, stateful optimization process, relying on runtime signals (e.g., $\nabla_{\mu_{j}}\mathcal{L}$ and $s_{j}$) to trigger cloning or splitting. This mechanism is fundamentally incompatible with our single-pass, feed-forward architecture, which lacks access to these iteratively accumulated, state-dependent signals.

This necessitates a principled, equivalent substitute that can be executed in a single forward pass, driven only by the per-point feature $f_{j}$. We propose Equivalent Densification Mechanism that simultaneously provides an equivalent substitute for both cloning and pruning. For cloning, the decoder $\mathcal{D}_{Gen}$ is trained to predict a fixed set of $K$ new primitives based on the single anchor feature $f_{j}$, effectuating a 1-to-K mapping:
\begin{equation}
\{\Delta\mu_{j,k}, c_{j,k}^{\prime}, s_{j,k}^{\prime}, R_{j,k}^{\prime}, \alpha_{j,k}^{\prime}\}_{k=1}^{K} = \mathcal{D}_{Gen}(f_{j}).
\end{equation}
For pruning, the network learns to implicitly remove redundant or incorrectly placed primitives by predicting a vanishing opacity ($\alpha_{j,k}^{\prime} \rightarrow 0$). This achieves a render-equivalent and fully differentiable substitute for the original hard pruning step.

Crucially, the final position $\mu_{j,k}^{\prime}$ of these new primitives is still determined by the Point Anchor Mechanism, ensuring our new density remains spatially consistent. This design allows the $\mathcal{D}_{Gen}$ module to learn an end-to-end function that directly approximates the ideal density distribution $\mathcal{D}(\mathcal{G}(T))$ in a single pass, efficiently replacing the complex iterative process.
\begin{equation}
\mathcal{D}(\mathcal{G}_{high}) = \mathcal{D}(\mathcal{F}_{\text{AnchorSplat}}(\mathcal{G}_{low})) \approx \mathcal{D}(\mathcal{G}(T)).
\end{equation}
The final high-quality asset $\mathcal{G}_{high}$ is composed of all newly generated primitives.

\subsection{Loss Function}

The training of the AnchorSplat model is supervised by high-resolution ground-truth images $I_{GT}$ and their camera poses, utilizing a low-quality 3DGS asset $\{\mathcal{G}_{\text{low}}\}$ as input. During the training process, the model generates $\mathcal{G}_{\text{HR}}$, from which a predicted image $\hat{I}$ is rendered and subsequently compared against $I_{GT}$. Gradients are back-propagated through the differentiable rendering pipeline to optimize the parameters of the AnchorSplat model.

The total loss function $\mathcal{L}_{\text{total}}$ is defined as a weighted summation of three critical components:
\begin{equation}
\mathcal{L}_{\text{total}} = \lambda_1 \mathcal{L}_{L1} + \lambda_2 \mathcal{L}_{\text{SSIM}} + \lambda_3 \mathcal{L}_{\text{Perception}},
\end{equation}
where $\lambda_1$, $\lambda_2$, $\lambda_3$ are balancing coefficients. $\mathcal{L}_{L1}$ is the mean absolute error between $\hat{I}$ and $I_{GT}$, which is employed to ensure basic pixel-wise similarity. $\mathcal{L}_{\text{SSIM}}$ is defined as $1 - \text{SSIM}(\hat{I}, I_{GT})$, serving as a perceptual metric to guarantee structural fidelity. $\mathcal{L}_{\text{Perception}}$ is a VGG-based perceptual loss that measures distance in the deep feature space, which is essential for maintaining the visual realism of both global appearance and fine-grained details.

\section{The 3DGS-SR Dataset}
\label{sec:dataset}

We position our work within the 3DSR domain, focusing on the asset enhancement paradigm. This paradigm is distinct from the traditional reconstruction from scratch approach, which optimizes a 3D representation from low-resolution images $\{I_{\text{LR}}\}$. Instead, asset enhancement aims to take an existing, low-quality 3DGS asset $\mathcal{G}_{\text{low}}$ as its direct input and enhance it into a high-fidelity counterpart. This common real-world task has historically faced a critical research bottleneck: the lack of a standardized, large-scale benchmark. To address this gap, we construct 3DGS-SR, a new benchmark comprising approximately 15k single-object assets sourced from Objaverse. It covers a wide diversity of categories (e.g., vehicles, furniture, household items) and designates 30 representative objects as a held-out test set, with the remainder used for training.

The core contribution of 3DGS-SR lies in its standardized $(\mathcal{G}_{\text{low}}, \{I_{\text{HR}}\})$ data pairs. To generate these, we first render 146 views for each asset from its pristine mesh at two resolutions: HR ground-truth images $\{I_{\text{HR}}\}$ at $1024 \times 1024$ and a corresponding LR images $\{I_{\text{LR}}\}$ at $256 \times 256$. This direct rendering process avoids the domain gap of 2D downsampling. Next, the task input $\mathcal{G}_{\text{low}}$ is generated by training the standard 3DGS algorithm on the LR images $\{I_{\text{LR}}\}$ for 15,000 iterations. Finally, we implement a PSNR-based screening mechanism, filtering out any asset where $\mathcal{G}_{\text{low}}$ achieves a PSNR below 34 against its own LR training images, ensuring a stable geometric baseline for all inputs. The evaluation protocol follows the original 3DGS methodology, where models enhance $\mathcal{G}_{\text{low}}$ and are then rendered from held-out test views for comparison against the $\{I_{\text{HR}}\}$ ground-truth.

\begin{figure*}[tbp]
\centerline{\includegraphics[width=\textwidth]{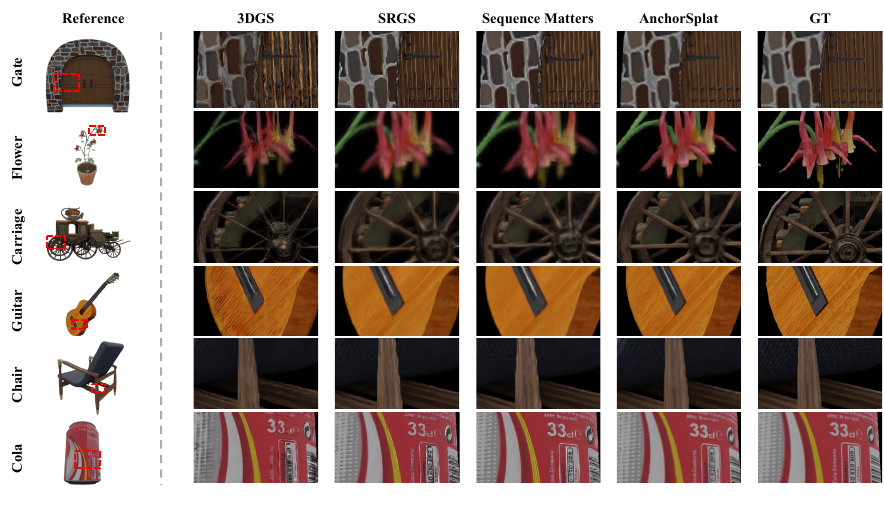}}
\caption{
Qualitative comparison on 3DGS-SR. We select several representative scenes from the 3DGS-SR test set for visual analysis. For image-based methods, the input is the LR image. 
2D super-resolution methods yield artifacts and oversmoothing at geometric edges and high-frequency details due to inconsistency; AnchorSplat avoids these phenomena.
}
\label{fig:model_compare}
\end{figure*}

\section{Experiments}

\subsection{Experimental Setting}
\label{sec:exp_setting}

Experiments are primarily conducted on the newly constructed 3DGS-SR benchmark, with zero-shot generalization evaluated on the NeRF-synthetic dataset. AnchorSplat is compared against representative 3DSR methods including Bicubic Optimization, SRGS~\cite{feng2024srgs}, SuperGaussian~\cite{shen2024supergaussian}, and SequenceMatters~\cite{ko2025sequence}. To ensure a fair comparison, all methods utilize identical low-resolution source information from the 3DGS-SR dataset. Specifically, reconstruction-based methods utilize LR images $\{ \mathcal{I}_{\text{LR}} \}$, while AnchorSplat takes the corresponding low-quality 3DGS assets $\mathcal{G}_{\text{low}}$ as input. 

Standard NVS metrics including PSNR, SSIM, and LPIPS are employed to evaluate rendering quality, with processing time (Time) reported to contrast the efficiency of the feed-forward paradigm against optimization-based methods. The model is implemented in PyTorch and trained via the Adam optimizer. The objective function $\mathcal{L}_{\text{total}}$ is defined as a weighted sum of $\mathcal{L}_{\text{L1}}$, $\mathcal{L}_{\text{SSIM}}$ (formulated as $1 - \text{SSIM}$), and a VGG-based $\mathcal{L}_{\text{Perception}}$. Training was conducted on 8 A800 GPUs.

\subsection{Results and Analysis}

\begin{table}[tbp]
\caption{Quantitative comparison on the 3DGS-SR dataset ($1024 \times 1024$). \textbf{Source-Free} indicates methods operating strictly on the 3D asset without accessing original multi-view images. AnchorSplat achieves state-of-the-art fidelity with real-time processing speed.}
\label{tab:gaussian_sr_results}
\centering
\setlength{\tabcolsep}{2.5mm}{
\begin{tabular}{l|c|ccc|c}
    \toprule[1pt]
    Method & Source-Free & PSNR$\uparrow$ & SSIM$\uparrow$ & LPIPS$\downarrow$ & Time$\downarrow$ \\
    \midrule[0.5pt]
    3DGS~\cite{kerbl20233d}& -- & 31.03 & 0.917 & 0.076 & -- \\
    \midrule[0.5pt]
    Bicubic & $\times$ & 34.36 & 0.923 & 0.064 & $\sim$ 4m \\
    SRGS~\cite{feng2024srgs} & $\times$ & 35.24 & 0.941 & 0.104 & $\sim$ 16m \\
    Sequence Matters~\cite{ko2025sequence} & $\times$ & 35.69 & 0.937 & 0.074 & $\sim$ 34m \\
    \midrule[0.5pt]
    SuperGaussian~\cite{shen2024supergaussian} & \checkmark & 34.94 & 0.924 & 0.097 & $\sim$ 41m \\
    AnchorSplat (Ours) & \checkmark & \textbf{36.57} & \textbf{0.943} & \textbf{0.058} & \textbf{$\sim$ 0.01s} \\
    \bottomrule[1pt]
\end{tabular}
}
\end{table}

\subsubsection{Main Results on 3DGS-SR}
Our primary objective is to enhance an existing low-quality 3DGS asset. Therefore, our most direct baseline is the input to AnchorSplat, represented by the 3DGS row in \cref{tab:gaussian_sr_results}, which is the low-quality asset reconstructed from LR images.

The quantitative results in \cref{tab:gaussian_sr_results} demonstrate that AnchorSplat achieves state-of-the-art performance, significantly surpassing the input baseline and all compared 3DSR paradigms, such as SRGS and SequenceMatters. This empirical evidence validates the two primary limitations of 2D-centric approaches: extremely low throughput and multi-view inconsistency. \cref{tab:gaussian_sr_results} clearly quantifies the critical efficiency bottleneck: while traditional optimization-based methods  require 0.25 to 1 hour of per-scene optimization, our single-pass, feed-forward network achieves this superior state-of-the-art quality in approximately 0.01 seconds. This realizes a $10^5$-fold increase in throughput.

Furthermore, these quantitative results are corroborated by the qualitative comparisons in \cref{fig:model_compare}. This visual evidence confirms that 2D-based methods visibly introduce the artifacts and geometric inconsistencies predicted in our introduction. In contrast, AnchorSplat successfully synthesizes fine, structurally-consistent details. This demonstrates that our 3D native approach is superior in both efficiency and fidelity, as the qualitative advantages are directly reflected in the state-of-the-art quantitative metrics.

\subsubsection{Generalization Ability and Efficiency Analysis}

The generalization capability of AnchorSplat is further evaluated by applying the model directly to the NeRF-synthetic dataset without fine-tuning. As shown in \cref{tab:blender_results}, AnchorSplat delivers substantial quality gains over baseline 3DGS assets, demonstrating robust cross-dataset performance.

\begin{table}[tbp]
\caption{Quantitative comparison on the NeRF-synthetic~\cite{mildenhall2021nerf} dataset. \textbf{Source-Free} indicates methods operating strictly on the 3D asset without accessing original multi-view images. AnchorSplat achieves the highest PSNR among all source-free methods while being orders of magnitude faster.}
\label{tab:blender_results}
\centering
\setlength{\tabcolsep}{2.5mm}{
\begin{tabular}{l|c|ccc|c}
    \toprule[1pt]
    Method & Source-Free & PSNR$\uparrow$ & SSIM$\uparrow$ & LPIPS$\downarrow$ & Time$\downarrow$ \\
    \midrule[0.5pt]
    3DGS~\cite{kerbl20233d} & -- & 23.30 & 0.872 & 0.114 & -- \\
    \midrule[0.5pt]
    Bicubic & $\times$ & 27.56 & 0.915 & 0.104 & $\sim$ 4m \\
    DiSR-NeRF~\cite{lee2024disr} & $\times$ & 26.00 & 0.889 & 0.122 & $\sim$ 30m \\
    NeRF-SR~\cite{wang2022nerf} & $\times$ & 28.46 & 0.921 & 0.076 & $\sim$ 24h \\
    Gaussian-SR~\cite{yu2024gaussiansr} & $\times$ & 28.37 & 0.924 & 0.087 & $\sim$ 15m \\
    SRGS~\cite{feng2024srgs} & $\times$ & 30.83 & 0.948 & 0.056 & $\sim$ 18m \\
    Sequence Matters~\cite{ko2025sequence} & $\times$ & \textbf{31.41} & \textbf{0.952} & \textbf{0.054} & $\sim$ 40m \\
    \midrule[0.5pt]
    SuperGaussian~\cite{shen2024supergaussian} & \checkmark & 28.44 & 0.945 & 0.067 & $\sim$ 45m \\
    AnchorSplat (Ours) & \checkmark & 28.97 & 0.935 & 0.077 & \textbf{$\sim$ 0.01s} \\
    \bottomrule[1pt]
\end{tabular}
}
\end{table}

This cross-dataset test also highlights the critical trade-off between fidelity and practicality under the Source-Free constraint. In comparison, methods such as Sequence Matters achieve a higher final PSNR; however, they require access to the original multi-view images to execute a complex pipeline involving 2D super-resolution and subsequent 3D re-optimization, costing approximately 40 minutes per scene. 

In stark contrast, AnchorSplat operates in a strictly Source-Free manner, achieving the highest PSNR among all methods that do not require original source images. Our method provides this substantial quality improvement in approximately 0.01 seconds, completely eliminating the need for any per-scene optimization or external image dependencies.

These results fully substantiate the powerful generalization of our model and its superior efficiency-accuracy trade-off. By outperforming other Source-Free alternatives like SuperGaussian in both quality and speed, AnchorSplat positions itself as a far more practical solution for high-fidelity deployment than conventional, time-consuming optimization methods.

\subsection{Ablation Study}

\begin{figure*}[tbp]
\centerline{\includegraphics[width=\textwidth]{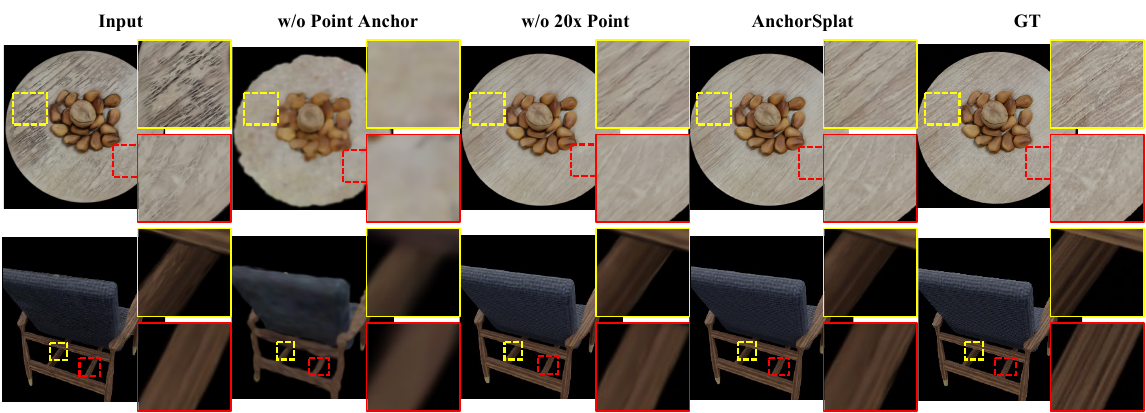}}
\caption{
Ablation study visualization. We perform separate ablations on the Point Anchor Mechanism and the Multiplicative Primitive Factor $K$. The visual results clearly demonstrate that the Point Anchor Mechanism is crucial for this task, while the generation of a high Multiplicative Primitive Factor ensures smoother and richer texture details.
}
\label{fig:ablation}
\end{figure*}

\begin{table}[tbp]
\caption{Ablation study on the 3DGS-SR dataset. Bold indicates the best result. The results demonstrate that our model achieves the best performance.}
\label{tab:ablation_results}
\centering
\setlength{\tabcolsep}{3mm}{
\begin{tabular}{c|c c c}
    \toprule[1pt]
    Method  &   PSNR$\uparrow$    &   SSIM$\uparrow$    &   LPIPS$\downarrow$  \\ \midrule[0.5pt]
    w/o Point Anchor     &   26.79   &   0.886   &   0.186  \\ 
    
    w/ 1$\times$  Points   &   36.42   &  0.943   &   0.063   \\ 
    w/ 10$\times$ Points   &   36.51   &   0.944   &  0.060   \\ 

    \midrule[0.5pt]
    Ours (20$\times$ Points)    & \textbf{36.57} &   \textbf{0.944}     & \textbf{0.058}   \\

    \bottomrule[1pt]

\end{tabular}
}
\end{table}

\subsubsection{Effectiveness of the Point Anchor Mechanism}
To validate the necessity of the Point Anchor Mechanism, we ablate this component and retrain the model. As shown in \cref{tab:ablation_results}, removing this module leads to a severe collapse in model performance. Quantitatively, all metrics sharply deteriorate: PSNR drops by nearly 10 dB, while SSIM falls significantly and the perceptual metric LPIPS indicates a severe degradation.

Qualitatively, \cref{fig:ablation} provides more direct visual evidence. Without the Point Anchor constraint, the rendered geometry and texture become chaotic and disorderly, with a near-total loss of high-frequency details. This result confirms that the Point Anchor Mechanism is fundamental. Without this local geometric constraint, the model suffers from the gradient confounding and mapping ambiguity issues inherent in direct 3D prediction, leading to ineffective optimization signals that prevent the learning of high-frequency information and result in the observed failure in detail synthesis.

\subsubsection{Effectiveness of Equivalent Densification Mechanism}
Next, we validate the contribution of primitive density, regulated by the Equivalent Densification Mechanism, to model accuracy. 

As shown in \cref{tab:ablation_results}, we compared the full model using a multiplicative factor of $K=20$ against variants with lower factors, such as $K=1$ and $K=10$. The results show that increasing the multiplicative factor from $K=1$ to $K=20$ consistently enhances all metrics, with PSNR improving and LPIPS dropping at each step.

\begin{figure*}[tbp]
\centerline{\includegraphics[width=\textwidth]{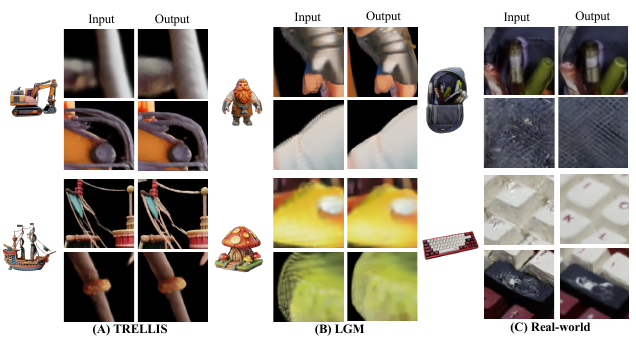}}
\caption{
Zero-shot generalization on diverse 3D sources. Without any fine-tuning, AnchorSplat acts as a plug-and-play enhancer for 3D generative models. It successfully sharpens mechanical boundaries for Trellis outputs (A), enriches complex geometric textures for LGM outputs (B), and robustly enhances details in unconstrained, noisy real-world captures (C).
}
\label{fig:generalization}
\end{figure*}

\cref{fig:ablation} visually confirms this finding. The $K=1$ low-density variant renders noticeably blurrier texture details and struggles to produce sharp high-frequency textures compared to the full model with $K=20$. This experimental observation confirms the direct positive impact of primitive densification on model accuracy, reflecting the necessity of a higher density of geometric primitives for synthesizing high-frequency details. 

This validates Equivalent Densification Mechanism as an effective and necessary component for achieving high-fidelity detail synthesis, confirming it serves as a successful substitute for traditional iterative densification.

\subsection{Generalization to 3D Generative Models and Real-World Scans}
\label{sec:generalization_aigc}

AnchorSplat demonstrates robust zero-shot generalization across diverse data distributions, ranging from feed-forward generative outputs of Trellis and LGM to noisy real-world captures. Although trained exclusively on synthetic datasets, our model consistently refines coarse geometry and restores high-frequency appearance details within a single 0.01s forward pass. This capability directly addresses the textural blurriness and structural ambiguity commonly observed in current large-scale 3D generative models, making AnchorSplat suitable as a lightweight post-processing module for generated 3D assets.

As illustrated in \cref{fig:generalization} (A), AnchorSplat significantly enhances the structural definition of Trellis outputs, particularly for mechanical geometries. For the excavator and ship models, our method recovers sharp structural boundaries and clarifies thin, intricate components that were initially rendered with substantial ambiguity. In LGM-generated assets shown in \cref{fig:generalization} (B), the Point Anchor Mechanism effectively suppresses generative artifacts and sharpens complex surface textures, such as the detailed armor of the dwarf and the organic patterns of the mushroom house. These results suggest that AnchorSplat does not merely amplify local colors, but learns to reorganize Gaussian attributes and positions in a structure-aware manner.

Crucially, this performance is achieved across a substantial domain gap in appearance modeling. While our model is trained on assets with a Spherical Harmonics (SH) degree of 3, it seamlessly processes the diffuse-only (SH=0) outputs of current generative pipelines without fine-tuning. This capability indicates that the model learns intrinsic geometric and structural priors rather than overfitting to a specific lighting representation. Furthermore, evaluations on unconstrained real-world scans presented in \cref{fig:generalization} (C), including objects such as the pencil case and keyboard, show that AnchorSplat remains resilient to capture noise, imperfect reconstruction, and sensor artifacts. By bridging coarse generative outputs and high-fidelity 3D assets, AnchorSplat provides a practical, plug-and-play solution for broader 3D content creation workflows.

\section{Conclusion}
\label{sec:conclusion}

This paper presents AnchorSplat, a 3D-native feed-forward network addressing low throughput and multi-view inconsistency in 3DGS enhancement. The proposed Point Anchor Mechanism mitigates ill-posed mapping, while the single-pass Equivalent Densification Mechanism replaces iterative optimization. Notably, AnchorSplat is a strictly source-free solution requiring no original multi-view images. We also introduce 3DGS-SR, the first large-scale benchmark for this task. Experiments demonstrate state-of-the-art fidelity with $10^{5}$ times throughput gains over optimization methods. Furthermore, AnchorSplat exhibits robust zero-shot generalization across generative outputs and real-world scans, establishing the framework as a practical enhancer for high-fidelity 3D synthesis.

\section*{Acknowledgements}

This work was supported by the New Generation Artificial Intelligence--National Science and Technology Major Project under Grant 2025ZD0123505; the National Natural Science Foundation of China under Grants 62576338, 62576342, 62506362, 62550062, 62425606, and 32341009; and the Beijing Natural Science Foundation under Grants L252145, L257008, and 4252054.

%
%
\bibliographystyle{splncs04}
\bibliography{main}
\end{document}